\documentclass{article}

\usepackage{microtype}
\usepackage{graphicx}
\usepackage{subfigure}
\usepackage{booktabs} 

\usepackage{hyperref}
\usepackage{natbib}

\usepackage{xcolor}
\definecolor{MyURL}{RGB}{0,0,139}       

\AtBeginDocument{%
  \hypersetup{
    linkcolor=MyURL,
    citecolor=MyURL,
    urlcolor=MyURL
  }%
}
\usepackage{multirow}

\usepackage[accepted]{mlsys2026}

\mlsystitlerunning{From Tokens to Layers: Redefining Stall-Free Scheduling for MoE Serving with Layered Prefill}

\begin{document}

\twocolumn[
\mlsystitle{From Tokens to Layers: Redefining Stall-Free Scheduling for MoE Serving with Layered Prefill}




\begin{mlsysauthorlist}
\mlsysauthor{Gunjun Lee}{snu}
\mlsysauthor{Jiwon Kim}{snu}
\mlsysauthor{Jaiyoung Park}{snu}
\mlsysauthor{Younjoo Lee}{snu}
\mlsysauthor{Jung Ho Ahn}{snu}
\end{mlsysauthorlist}

\mlsysaffiliation{snu}{Seoul National University, Seoul, South Korea}

\mlsyscorrespondingauthor{Gunjun Lee}{kevin970401@snu.ac.kr}
\mlsyscorrespondingauthor{Jung Ho Ahn}{gajh@snu.ac.kr}

\mlsyskeywords{Machine Learning, LLM inference, LLM serving}

\vskip 0.3in

\begin{abstract}
Large Language Model (LLM) inference in production must meet stringent service-level objectives for both time-to-first-token (TTFT) and time-between-token (TBT) while maximizing throughput under fixed compute, memory, and interconnect budgets. Modern serving systems adopt stall-free scheduling techniques such as chunked prefill, which splits the processing of long prompts along the token dimension and interleaves prefill with ongoing decode iterations. While effective at stabilizing TBT, chunked prefill incurs substantial overhead in Mixture-of-Experts (MoE) models: redundant expert weight loads increase memory traffic by up to \textbf{39\%} and inflate energy consumption. We propose \textbf{layered prefill}, a new scheduling paradigm that treats transformer layer groups as the primary scheduling unit, specifically targeting MoE serving. By vertically partitioning the model into contiguous layer groups and interleaving prefill and decode across the groups, layered prefill sustains stall-free decoding while eliminating chunk-induced MoE weight reloads. It reduces off-chip bandwidth demand, lowering TTFT by up to \textbf{70\%}, end-to-end latency by \textbf{41\%} and per-token energy by up to \textbf{22\%}. Evaluations show that layered prefill consistently improves the TTFT--TBT Pareto frontier over chunked prefill, reducing expert-load traffic and energy cost while maintaining stall-free decoding. Overall, shifting the scheduling axis from tokens to layers unlocks a new operating regime for high-efficiency, energy-aware MoE serving in co-located environments.
\end{abstract}
]



\printAffiliationsAndNotice{}  

\section{Introduction}

Transformer-based~\cite{neurips-2017-attention} large language models (LLMs) have demonstrated remarkable performance across domains once thought to require exclusively human capabilities, including question answering, code generation, and mathematical problem solving~\cite{arxiv-2025-deepseek_r1, arxiv-2025-qwen3, arxiv-2025-gptoss}. Beyond solving discrete tasks, they are now widely used to assist or automate work ranging from complex reasoning to repetitive operations. Rapid adoption of LLMs has led cloud providers to build massive accelerator clusters (AI factories), and has simultaneously sparked intensive systems research on LLM serving~\cite{arxiv-2022-efficient_scaling, osdi-2024-distserve, arxiv-2025-beyondthebuzz, arxiv-2025-bottleneck, osdi-2022-orca, osdi-2024-sarathi}.

LLM performance improves consistently with model size, and \textbf{Mixture-of-Experts (MoE)} has emerged as the dominant strategy for scaling. Instead of using a single feed-forward network (FFN) block, an MoE architecture employs multiple experts and a routing mechanism that assigns each token to a small subset of experts. Because only a fraction of experts are activated per token, the number of arithmetic operations per token remains comparable to a standard FFN block, while the total parameter count can grow dramatically. This sparse activation enables models to scale to trillions of parameters, which in turn translates into consistent improvements in downstream performance~\cite{iclr-2017-moe, iclr-2021-gshard, jmlr-2022-switch}.

LLM inference is governed by two stages that stress different hardware bottlenecks. The prefill stage processes the entire input sequence at once to create the key-value (KV) cache and generate the first output token. Since LLM weights are reused over all tokens of an input prompt, this stage is typically compute-bound. The decode stage then forwards the previous token, updates the KV cache, and generates the next output token, repeating the process one token at a time. This stage is typically memory-bound as it involves loading the key-value cache and other parameters for each token being generated.
Consequently, batching is highly effective for decode, where weight reuse improves throughput, but provides little benefit for prefill, creating an inherent asymmetry in LLM serving.

LLM serving performance is primarily characterized by two latency metrics: time to first token (TTFT) and time between tokens (TBT). TTFT measures the total time from a request's arrival to the generation of its first token, encompassing both the queuing and prefill phases.
TBT is 
the time interval between subsequent output token generations during the decode phase, which reflects how well the system can sustain token generation once the first token has been produced.
Service level objectives (SLOs) define the per-request latency budgets that a serving system must meet; for our purposes, the TTFT SLO bounds first-token latency and the TBT SLO bounds each inter-token interval.

The dual goals of maintaining stable TTFT and TBT are often in conflict. If prefill work cannot be sufficiently overlapped with decode, the additional work inflates per-step latency and triggers TBT violations. This constraint forces a trade-off: either tolerate higher TTFT (e.g., by deferring or batching prefill) or sacrifice throughput (e.g., by reducing concurrency or batch size). Thus, the core challenge in LLM serving is to push the Pareto frontier among TTFT, TBT, and throughput under fixed hardware resources.

Early serving systems such as FasterTransformer~\cite{github-2019-fasttransformer} adopted a straightforward approach: process requests from start to finish in fixed batches. This guarantees stall-free decoding and stable TBT, but new requests must wait until the current batch completes, inflating TTFT. More fine-grained schedulers, such as those in Orca~\cite{osdi-2022-orca} and vLLM~\cite{sosp-2023-vllm}, moved to iteration-level scheduling. Specifically, Orca's continuous batching inserts prefill of new requests immediately after each decode iteration, reducing TTFT while preserving throughput via a steady pool of active decode work.

The shift towards long prompts, however, exposed a new bottleneck: decode batches often stall while waiting for a single, large prefill to finish, triggering TBT SLO violations.
Chunked prefill~\cite{osdi-2024-sarathi} addresses this by splitting long prompts into smaller chunks and interleaving their prefills with decode.
The \emph{chunk size} (number of tokens per chunk) is the key tuning knob: smaller chunks reduce per-iteration prefill work and relax TBT pressure, whereas larger chunks reduce the number of chunks and lower TTFT.
By keeping per-iteration prefill work within the available slack, chunked prefill suppresses TBT spikes and restores stall-free execution.

However, this benefit of chunking comes with a structural cost in MoE models.
Partitioning the input along the token dimension forces each chunk to traverse the same transformer layers and repeatedly reload expert weights, amplifying expert-load counts; the problem is exacerbated as chunk size shrinks to meet tight TBT SLOs. In contrast, scheduling on layer boundaries aligns with MoE expert load units and eliminates redundant KV-cache scans and weight loads, while still allowing stall-free execution.

This paper proposes \textbf{layered prefill}, which treats the layer dimension rather than the prompt (token) dimension as the primary scheduling resource, specifically targeting MoE serving. The decoder stack is divided into $N_{lg}$ contiguous \emph{layer groups} (contiguous subsets of transformer layers); $N_{lg}$, the \emph{number of layer groups}, is the key tuning knob of layered prefill, playing the same role as chunk size in chunked prefill. In each iteration, prefill is performed alongside decode for exactly one group, while the remaining groups perform decode only. This design ensures that prefill completes in exactly $N_{lg}$ iterations without stalling decode. During prefill, each layer is traversed exactly once, which fundamentally avoids memory-access amplification from repeatedly loading experts.

Our key contributions are as follows:
\begin{itemize}
    \item We treat layers as a first-class scheduling unit and enforce a one-group-per-iteration rule that preserves stall-free decoding for MoE models.
    \item We eliminate chunk-amplified MoE weight reloads through layered prefill, systematically reducing redundant parameter traffic and improving the TTFT–TBT trade-off in long-context, small-batch settings.
    \item  We reduce energy consumption during MoE serving by cutting redundant MoE reloads, lowering off-chip parameter movement, and decreasing GPU DRAM activity.
\end{itemize}

\section{Large Language Model (LLM)}
\subsection{LLM Architecture}
Contemporary LLMs such as GPT~\cite{neurips-2020-gpt3, arxiv-2024-gpt-4}, Llama~\cite{arxiv-2024-llama3}, and Qwen~\cite{arxiv-2025-qwen3} typically adopt a decoder-only Transformer architecture~\cite{neurips-2017-attention}.
A model maps a sequence of token IDs into continuous vector representations, processes them through a stack of Transformer decoder blocks, and finally projects the resulting hidden states into vocabulary logits to predict the next tokens.

Each decoder block follows the same structure: a self-attention layer followed by a feed-forward network (FFN), with residual connections and normalization applied around both.
Self-attention enables each token (represented as a vector) to integrate contextual information from earlier tokens in the sequence.
It projects input hidden states to query, key, and value matrices via learned projection weights ($W_Q$, $W_K$, $W_V$), applies a causal mask to ensure that tokens attend only to past and current positions, and derives attention weights through softmax-normalized dot products between queries and keys.
These weights guide a weighted sum over the values, producing a context-aware token representation.
A rich line of work accelerates the attention operator---for example, FlashAttention and its successors~\cite{neurips-2022-flashattenion, iclr-2024-flashattention_v2, neurips-2024-flashattention_v3}---which reduce attention I/O and improve kernel efficiency on modern AI accelerators, such as GPUs~\cite{micro-2023-nvidia, hcs-2024-nvidia} and TPUs~\cite{isca-2023-tpu, micro-2021-design}.

FFN refines each token's hidden state independently.
It consists of two or more fully connected (FC) layers with non-linear activations, such as ReLU or SwiGLU~\cite{arxiv-2020-glu}.
Conventionally, the first FC layer expands the hidden dimension by a factor of 2--4 to increase representational capacity, while the last layer projects it back to the original size to preserve dimensional consistency across layers.

\subsection{LLM Inference Structure: Prefill and Decode}
Standard LLM inference---unless using diffusion or similar methods---is autoregressive; it generates output tokens one by one while depending on previously generated tokens.
The process consists of multiple passes through a full stack of decoder blocks, referred to as iterations.
These iterations are grouped into two distinct stages: \emph{prefill} and \emph{decode}.

The prefill stage corresponds to the first iteration.
It processes the entire sequence of prompt tokens in parallel along the sequence dimension, passing all tokens through every decoder block.
This stage achieves the high compute utilization of AI accelerators due to its parallelism~\cite{arxiv-2025-prefillonly}; however, it also incurs high latency as all prompt tokens must be processed before producing the first output token (TTFT).

The decode stage consists of all subsequent iterations, each generating a new token per request.
As the sequence length per iteration is only one token, compute units are underutilized~\cite{asplos-2024-attacc}, and its efficiency depends heavily on memory access patterns.
To improve utilization, systems typically batch decode iterations from multiple requests.

\subsection{Prior LLM Serving Approaches and Challenges}

\begin{figure*}[tb!]
    \centering
    \includegraphics[width=0.99\textwidth]{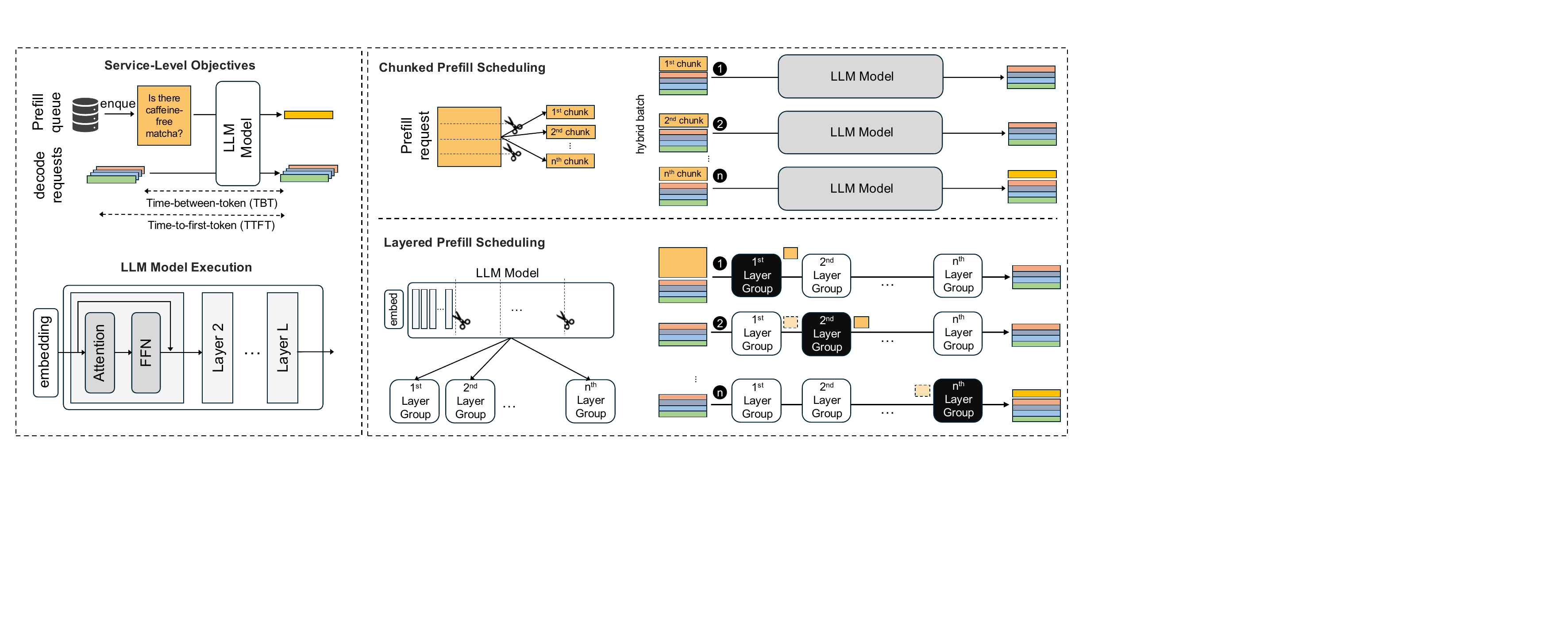}
    \vspace{-0.03in}
    \caption{
    (Upper right) Per iteration, chunked prefill splits an input prompt into multiple chunks, and at each iteration one chunk is processed in order from the beginning with the decode.
    (Lower right) For layered prefill, exactly one layer group performs both prefill and decode, while the others perform decode only.
    Prefill advances by one group per iteration, maintaining stall-free decoding.}
    \label{fig:scheduling}
\end{figure*}

LLM serving systems have evolved rapidly to sustain growing request volumes under strict latency service level objectives (SLOs).
A series of recent works~\cite{osdi-2022-orca, sosp-2023-vllm, osdi-2024-sarathi, osdi-2024-distserve, isca-2024-splitwise, arxiv-2025-beyondthebuzz, asplos-2025-podattention, arxiv-2025-pdagg} proposed increasingly sophisticated scheduling policies aimed at improving utilization while preserving responsiveness.

Early systems such as FasterTransformer~\cite{github-2019-fasttransformer} relied on static or per-request batching: each batch was fixed and processed independently, without dynamically merging arriving requests. While straightforward to implement, this design yields poor hardware utilization in the decode stage, where only a small number of vectors are processed concurrently, and decode batches can stall behind long prefills, causing high time-between-token (TBT) latency for concurrent requests.

To mitigate these inefficiencies, systems such as Orca~\cite{osdi-2022-orca} introduced continuous batching, which merges prefill and decode requests and schedules them together at the iteration level, substantially improving inference throughput.
However, this approach still suffers from stalls when long prefills delay subsequent decode steps.

Sarathi-Serve~\cite{osdi-2024-sarathi} addresses this issue through chunked prefill.
Instead of processing an entire long input in one pass, the sequence is split into smaller chunks that are interleaved with decode requests to form \emph{hybrid} batches.
This restructuring enables stall-free\footnote{We define stall-free scheduling as the state where, for each request, every TBT does not exceed a model-specific threshold.} execution, since decode requests are no longer blocked behind lengthy prefills.
Further, by setting the chunk size---typically 256 or 512 tokens---the scheduler both satisfies the TBT constraint and distributes prefill work more evenly across iterations, which improves hardware utilization.

As depicted in Figure~\ref{fig:scheduling} (left), chunked prefill processes one chunk per iteration, processing all decoder layers of the current chunk before moving to the next one.
Each chunk reuses the KV cache from preceding chunks, continuing until the full prompt is processed.
To improve efficiency, short requests may be coalesced into a single chunk.
The chunk size is capped to meet tail-latency SLOs, limiting per-iteration batch size.
Consequently, chunked prefill significantly improves service-level objectives (e.g., TTFT and TBT), and we adopt it as the baseline in our evaluation.

\subsection{Recent Trends in LLM Architecture and Usage}

\noindent\textbf{Mixture of Experts (MoE):}
MoE architectures~\cite{iclr-2017-moe, iclr-2021-gshard, jmlr-2022-switch} have emerged as a promising approach to scaling model size without proportionally increasing computational complexity.
Unlike dense FFNs, which activate all parameters even with small decode batches, MoE routes tokens to a small subset of independent FFNs, referred to as experts, via routers.
This sparsity efficiently scales parameter count by activating only a fraction of the model’s parameters per token; however, it also introduces challenges in scheduling and memory access~\cite{arxiv-2022-stmoe, icml-2022-deepspeed_moe}.

\noindent\textbf{Long Context:}
Recent LLM workloads involve processing much longer input prompts than in the past.
Prior systems typically handled prompts in hundreds of tokens~\cite{neurips-2020-gpt3}, whereas modern applications driven by multi-turn conversations, retrieval-augmented generation, and chain-of-thought reasoning require contexts of tens of thousands of tokens.
This trend is accelerated by the deployment of long-context LLMs such as Claude~3~\cite{anthropic-2024-claude3}, Gemini~2.5~\cite{arxiv-2025-gemini2_5}, GPT-OSS~\cite{arxiv-2025-gptoss}, and DeepSeek-R1~\cite{arxiv-2025-deepseek_r1}, exceeding a 100k-token context window.
Consequently, the prefill stage has become more computationally expensive and memory-intensive, exacerbating the bottlenecks in contemporary, co-located serving systems where both prefill and decode are processed by the same AI accelerators.

These advances in LLM architecture and workloads both increase the computational and memory demands of LLM serving, underscoring the need for new scheduling and parallelism techniques to maintain SLO compliance.

\subsection{Energy Consumption in LLM Serving}
Power and energy efficiencies are first-order concerns for large-scale LLM serving.
Beyond meeting latency-oriented SLOs, service providers must minimize the energy required to serve requests for saving both operating cost and carbon footprint~\cite{hpca-2025-dynamollm}.

\noindent\textbf{Energy Breakdown and Accounting:}
The energy cost of LLM serving on AI accelerators can primarily be viewed as the sum of four components: (i) a static baseline that keeps the device active even when idle, (ii) compute energy dominated by matrix–multiply/accumulate engines (e.g., Tensor Cores in GPUs~\cite{pds-2020-gpu, icpe-2024-accelerating} and MXUs in TPUs~\cite{tech4future-2023-tensor, ccgrid-2020-benchmarking}), (iii) memory energy from moving parameters and KV cache through the memory hierarchy (on-chip SRAM and off-chip DRAM, such as HBM or GDDR), and (iv) communication energy from interconnects that link devices or hosts (e.g., NVLink, PCIe, InfiniBand, and Ethernet).

In practice, data movement accounts for a large share of energy, since each generated token repeatedly rereads substantial portions of the model parameters.
This observation can be verified by comparing the power consumption of a $4$-bit quantized model (W4A16) with that of a higher-precision counterpart~\cite{arxiv-2025-quant}.
A practical accounting, thus, is to track the total bytes moved at each level of the memory hierarchy and multiply by the empirically measured energy-per-byte for that level, with the off-chip DRAM term typically setting the overall scale.

\noindent\textbf{MoE Implications:}
Whether the GEMM is compute-bound or memory-bound can be checked by comparing the batch-implied arithmetic intensity against the accelerator's ridge point. The ridge point is defined as peak arithmetic throughput divided by peak memory bandwidth; it marks the Op-per-Byte (Op/B) at which execution shifts from memory-bound to compute-bound. Operating near the ridge point means that the system utilizes both memory and compute efficiently. Modern AI accelerators typically have ridge points around 100 to 300 Op/B~\cite{arxiv-2025-bottleneck}. This implies a batch size of roughly 200 to 600 for a 2-byte datatype (e.g., bflot16).

With this in mind, although MoE activates only a subset of experts per token, whether this sparsity actually reduces latency and energy in a serving system requires careful analysis. Consider an MoE model with 128 experts and top-$k$ of 8 such as Qwen3-30B-A3B. For an input prompt of 2048 tokens, each expert on average processes about 128 tokens. This falls below the ridge point of contemporary accelerators, so each expert's FC compute remains memory-bound and end-to-end latency and energy are dominated by loading MoE weights from HBM. Making MoE computation compute-bound requires more than 8192 tokens, which would cause a sharp increase in latency (TBT).

\section{Limitations of Chunked Prefill on LLMs with MoE}

\subsection{MoE Weight Load Model}

Chunked prefill introduces a conflict for MoE layers.
To satisfy tail-latency SLOs, a chunked prefill scheduler limits per-iteration batch sizes (\S\ref{sec:e2e_slo}) whereas efficient MoE weight-loading requires much larger batches to maximize reusing loaded experts.
This mismatch makes expert weights broadly activated but poorly reused, leaving execution bottlenecked by memory traffic rather than computation.

Suppose the batch size is 128, top-$k$ is 8, and the total number of experts is 128; on average, only 8 tokens are routed to each expert per iteration.
Eight is far below the ridge point of modern AI accelerators.
Thus, we formulate this issue as a \emph{sparsity erosion} problem; chunking inflates expert coverage while suppressing reuse, eroding the very sparsity benefits that make MoE efficient.
In the following, we quantify expert activation under realistic serving conditions and analyze how chunking disrupts MoE’s sparsity advantages.

\begin{table}[tb!]
\centering
\caption{Expert weight coverage ratio as a function of decode batch size, measured on \texttt{Qwen} with the ShareGPT dataset.}
\label{tab:moe_weight_ratio}
\vskip 0.15in
\setlength{\tabcolsep}{3pt}        
\renewcommand{\arraystretch}{0.95} 
\resizebox{\columnwidth}{!}{%
\begin{tabular}{@{}lcccccccccc@{}}
\toprule
\textbf{Batch Size} & 1 & 2 & 4 & 8 & 16 & 32 & 64 & 128 & 256 & $\ge$512 \\
\midrule
\textbf{Coverage (\%)} & 6.25 & 11.7 & 21.3 & 29.0 & 44.5 & 54.7 & 69.4 & 86.3 & 93.4 & $\ge$98 \\
\bottomrule
\end{tabular}}
\end{table}

\begin{figure}[t!]
    \centering
    \includegraphics[width=0.96\columnwidth]{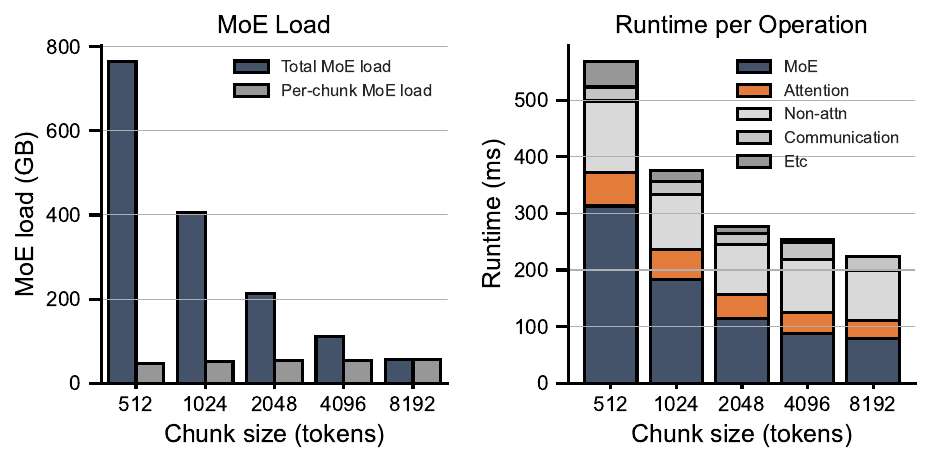}
    \vspace{-0.05in}
    \caption{(Left) MoE weight loading vs.\ chunk size.
    Per-chunk MoE load remains roughly constant regardless of chunk size, and the total MoE load equals the per-chunk MoE load multiplied by the number of chunks. Therefore, as chunk size increases, the total MoE load tends to decrease.
    (Right) Runtime of each kernel vs.\ chunk size, with operations broken down by category
    (MoE, attention, linear projections, communication, and etc.).
    The input length is fixed at 8{,}192 tokens.}
    \label{fig:duration_chunk}
\end{figure}

Measurements on Qwen3-30B-A3B~\cite{arxiv-2025-qwen3} (\texttt{Qwen}) evaluated on the ShareGPT dataset~\cite{sharegpt-_-sharegpt} showcase this issue  (see Table~\ref{tab:moe_weight_ratio}).
Details of the model and dataset are provided in Table~\ref{tab:moe_model_info} and Table~\ref{tab:dataset_stats}.
At small batch sizes (below 16), fewer than half of the experts are loaded, and even with a moderate batch size of 64, the average expert coverage remains under 70\%.
As shown in Figure~\ref{fig:slo_distribution}, in stable serving regimes (e.g., where SLO attainment exceeds 90\%), the average batch size is below 32 on the arXiv dataset~\cite{arxiv-2018-arxiv} and below 128 on the ShareGPT dataset, implying that only a limited fraction of experts is activated per decode batch.
In pure decoding, this limited activation is beneficial: it reduces memory traffic relative to dense models.

In contrast, chunked prefill creates hybrid batches that are typically much larger (the batch size often exceeding 256), which activates most experts and erodes the sparsity benefit.
At the same time, TBT SLO constraints cap the chunk size; thus, the number of tokens routed to each expert remains insufficient to saturate compute resources, leaving MoE layers memory-bound.
Therefore, memory traffic dominates system behavior, amplifying per-iteration latency and energy cost in proportion to prompt length and chunk count.

\subsection{Microbenchmark: Runtime vs. Chunk Size}
To understand how chunk size shapes MoE efficiency, we isolate its effect on kernel performance.
Figure~\ref{fig:duration_chunk} reports MoE weight loading and kernel runtimes as a function of prefill chunk size on \texttt{Qwen} (evaluation setup described in \S\ref{sec:experimental_setup}), with the input length fixed at 8{,}192 tokens to approximate the average prompt length in the arXiv dataset.

When the chunk size is 512, MoE runtime accounts for over 50\% of the prefill runtime, and the prefill runtime exceeds 500 ms. As chunk size increases, weight-loading overhead falls sharply, 
in inverse proportion to chunk size, and runtime per operation decreases correspondingly. Larger chunks route more tokens to each expert, improving reuse and reducing redundant transfers. By 4096--8192 tokens, MoE load drops below 100 GB and prefill runtime stabilizes near 200 ms, with MoE no longer dominating execution. Other kernels modestly depend on chunk size, largely due to lower arithmetic intensity at small problem sizes, but the effect is far less pronounced than in MoE kernels.

These results demonstrate that chunk size strongly governs MoE efficiency: small chunks trigger sparsity erosion and memory-bound slowdowns, whereas larger chunks mitigate the problem by enabling efficient expert reuse.

\subsection{Chunk Size Trade-offs in LLM Serving}

\begin{table}[tb!]
\centering
\caption{Chunk size trade-offs for \texttt{Qwen} on the arXiv workload. Larger chunks improve runtime and energy efficiency but inflate tail TBT.}
\label{tab:chunk_size_tradeoff}
\vskip 0.15in
\resizebox{\columnwidth}{!}{%
\begin{tabular}{ccccccccc}
\toprule
Chunk & Req. & \multicolumn{2}{c}{TTFT (s)} & \multicolumn{2}{c}{TBT (ms)} & Load & Energy \\
Size  & Rate & Mean & p99 & Mean & p99 & (GB/req) & (mJ/tok) \\
\midrule
512   & 1.3  & 2.68 & 8.05 & 29.0 & 48.4 & 955 & 60.2 \\
1024  & 1.7  & 2.32 & 5.83 & 43.6 & 83.4 & 631 & 45.4 \\
2048  & 2.6  & 2.56 & 5.58  & 73.6 & 129 & 304 & 32.4 \\
\bottomrule
\end{tabular}}
\end{table}

In a serving scenario, chunk size shapes three key aspects of system behavior: expert reuse, serving capacity, and iteration latency. Larger chunks improve MoE efficiency by increasing reuse and reducing redundant weight load, and they also sustain higher request rates. At the same time, however, larger chunks extend iteration duration, which inflates tail latency and risks violating SLOs.

Table~\ref{tab:chunk_size_tradeoff} summarizes this trade-off for serving \texttt{Qwen} on the arXiv workload.
Request rates for each chunk size are adjusted to keep TTFT approximately 2.5\,s.
As chunk size increases, prefill runtime decreases.
Thus, holding TTFT constant across chunk sizes implies that larger chunks can sustain higher request rates so that the sum of queuing delay and prefill runtime remains 2.5\,s. 
From an efficiency perspective, larger chunks reduce energy consumption per token---defined as total energy divided by the sum of prompt and generated tokens~\cite{euromlsys-2025-advocating, energy-2025-storellm}---from 60\,mJ/tok at 512 to 32\,mJ/tok at 2048 ($-46$\%).
This improvement arises as fewer chunks per request lower the expert weight reloads and increase reuse.
Throughput also improves with larger chunks; the system sustains 1.3\,req/s at a chunk size of 512 but increases to 2.6\,req/s at 2048, consistent with the runtime trends in Figure~\ref{fig:duration_chunk}.

However, this conflicts with Sarathi-Serve's original configuration, as large chunk sizes lead to violations of the TBT constraint. As chunk size increases, p99 TBT rises sharply from 48\,ms at 512 tokens to 129\,ms at 2048 tokens, exceeding the SLO.
It is a direct consequence of longer iteration times, which delay the scheduling of concurrent decode requests and inflate the tail of the distribution. Thus, the chunk size that appears optimal from the standpoint of efficiency and throughput is infeasible under strict latency requirements.

\section{Layered Prefill Design}

\subsection{Overview}

To address the aforementioned limitations, we propose \textbf{layered prefill}.
The key idea is to shift the scheduling axis from tokens to layers.
Instead of partitioning the input sequence into chunks along the token dimension, the model is vertically partitioned into contiguous \emph{layer groups} (see Figure~\ref{fig:scheduling}).
This group-based layer partitioning naturally aligns with decoder-only Transformers, where layers are homogeneous and executed sequentially.

During execution, layered prefill follows a \emph{one-group-per-iteration} scheduling rule: in each iteration, only one designated layer group performs both decode and prefill for newly admitted requests, while all other groups perform decode only. Therefore, prefill is distributed across iterations, while decoding proceeds continuously without stalls.

This design guarantees that each input prompt traverses the prefill path of each layer exactly once, as opposed to chunked prefill, which forces every layer to process the prompt once per chunk. Consequently, redundant expert-weight reloads across chunks are eliminated, which reduces memory-bandwidth consumption and energy while preserving stall-free decoding under strict TBT SLOs.

\subsection{Scheduling Mechanism}

At every iteration, up to a single designated layer group jointly processes the ongoing decode batch and the prefill of incoming requests; the remaining groups execute decode-only computation.
Thus, prefill progresses group by group across iterations, while decoding remains uninterrupted.

Figure~\ref{fig:scheduling} illustrates the process with four layer groups. In the first iteration, prefill for a new request runs within Group~1, whereas Groups~2--4 perform decode only.
In the next iteration, prefill advances to Group~2, co-scheduled with decode, and so on.
This process repeats until the request has traversed all groups.
After $N_{lg}$ (the number of groups) iterations, the prefill of that request completes, while decoding remains stall-free throughout.

Unlike chunked prefill---which induces redundant MoE weight reloads across chunks---the layered prefill schedule ensures that each layer processes the input prompt only once. Thus, decoding stays stall-free while avoiding the amplified memory overheads of token-level partitioning.

Decode is never blocked because every iteration includes decode work across all layer groups, so TBT remains low enough to meet the SLO constraint. \emph{In short, moving from chunked prefill to layered prefill preserves stall-free decoding while substantially reducing redundant MoE weight movement and energy consumption.}

\subsection{Generalization with Chunked Prefill}

Layered prefill divides the model vertically into $N_{lg}$ contiguous layer groups, whereas chunked prefill partitions the input prompt into multiple chunks along the token axis.
Thus, the two methods are orthogonal, meaning that chunked prefill and layered prefill can be applied simultaneously.

Applying both strategies together increases scheduling flexibility and provides two key benefits.
First, it enables the efficient processing of very long input sequences.
Chunked Pipeline Parallelism, proposed in~\cite{arxiv-2025-beyondthebuzz}, addresses the challenge of handling extremely long input lengths by partitioning the input into relatively small chunks and pipelining the prefill of each chunk.
This method has been shown to be an optimal approach to meeting TTFT constraints for long inputs without requiring complicated parallelism strategies.
Since layered prefill is also compatible with chunking, it inherits the advantages of chunked pipeline parallelism in processing long inputs.

Second, it allows the chunk size to be substantially increased.
Increasing the chunk size directly reduces the number of chunks for a given input prompt.
Because the MoE weight read overhead grows with the number of chunks, fewer chunks dramatically reduce memory bandwidth consumption and energy cost.
Moreover, when the chunk size becomes sufficiently large, the batch size per expert in MoE layers also becomes large enough to saturate compute, shifting the operation from a memory-bound to a compute-bound regime.
For example, in a model with 128 experts and top-$k=8$, a chunk size of 8192 already yields an effective batch size of 512 tokens per expert, which is sufficient to make the MoE layer compute-bound. Once the chunk size becomes large enough to shift the MoE layer into the compute-bound regime, an important advantage emerges: partitioning the input prompt into multiple chunks for MoE computation no longer leads to a surge in latency.

Through this generalization, the system can selectively combine chunked prefill and layered prefill, leveraging the strengths of both approaches while mitigating their individual drawbacks.

\begin{table}[t!]
\caption{Specifications of the evaluated MoE models, reorganized by metric. GQA stands for grouped-query attention.}
\label{tab:moe_model_info}
\vskip 0.1in
\begin{center}
\begin{small}
\begin{tabular}{r|c|c}
\toprule
\textbf{Metric} & \textbf{\texttt{Qwen}} & \textbf{\texttt{GPT}} \\
\midrule
  Parameters                 & 30B        & 20B \\
  Number of total experts    & 128        & 32 \\
  Top-$k$ (active experts)   & 8          & 4 \\
  Experts-to-top-$k$ ratio   & 16:1       & 8:1 \\
  KV cache size per token    & 48\,KB     & \textless{}34\,KB\footnotemark \\
  Attention                  & GQA        & GQA \\
  Hidden dimension           & 2048       & 2880 \\
\bottomrule
\end{tabular}
\end{small}
\end{center}
\end{table}

\footnotetext{\texttt{GPT} uses sliding window attention~\cite{arxiv-2020-longformer} so that cache size per token varies over the time.}

\subsection{Implementation Details}
\label{sec:implementation_details}

We adapt the number of layer groups to the input length. Concretely, we choose the number of groups $N_{lg}$ so that the \emph{prefill work per iteration} approximately matches a 512-token chunk in the chunked prefill baseline:
\[
N_{lg}(L) \;=\; \max\bigl(1,\ \lceil L/512 \rceil \bigr).
\]
Hence, for a long prompt of length 8{,}192 we set $N_{lg}=16$, whereas for a short prompt of length 512 we set $N_{lg}=1$.
This choice aligns the processing granularity with chunked prefill at a chunk size of 512, making iteration time and admission cadence comparable across schedulers and isolating the effect of switching the scheduling axis (layers vs. tokens).
Here, 512 is an arbitrary value and could just as well be 256 or 1024. We use 512 in our experiments to ensure a fair comparison with chunked prefill configured with a chunk size of 512.
When combining chunked prefill and layered prefill for very long contexts, we typically use a large chunk size (e.g., 8192) and apply layered prefill with a specific group size (e.g., $N_{lg}=16$) within each chunk.
When multiple small inputs arrive concurrently, we merge them into a single batch to improve efficiency.

We implement layered prefill on top of vLLM~\cite{sosp-2023-vllm}.
Attention leverages FlashAttention-3~\cite{neurips-2024-flashattention_v3}; most computations use custom CUDA kernels, and the remaining layers use \texttt{torch.compile}.
We adopt CUDA Graphs to accelerate the prefill and decode stages.

\section{Evaluation}

\subsection{Experimental Setup}
\label{sec:experimental_setup}

\noindent\textbf{Environment:}
We conducted all experiments on a server equipped with two NVIDIA H100 GPUs (80\,GB each) connected via NVLink.
We implemented model parallelism using tensor parallelism.

\noindent\textbf{Models:}
We evaluated two representative MoE models with similar architectures but different routing configurations.
We abbreviate Qwen3-30B-A3B as \texttt{Qwen} and GPT-OSS-20B as \texttt{GPT} for brevity.
\texttt{Qwen} comprises 128 experts, and each token activates top-$8$ experts.
\texttt{GPT} comprises 32 experts, and each token activates top-$4$ experts.
Detailed specifications are summarized in Table~\ref{tab:moe_model_info}.
Both employ bfloat16 precision for weights and activations.

\noindent\textbf{Datasets:}
To analyze the effectiveness of layered prefill, we evaluate on two datasets with distinct characteristics, summarized in Table~\ref{tab:dataset_stats}.
\emph{ShareGPT} contains multi-turn conversations between users and ChatGPT.  ShareGPT exhibits a wide spread in input lengths---from a few hundred tokens to more than 10{,}000-resulting in a large input---length standard deviation.
On average, its input prompts are roughly six times longer than their outputs.
\emph{arXiv Summarization} (hereafter, \emph{arXiv}) is a summarization benchmark in which inputs are substantially longer than outputs; on average, the input length is approximately forty times the output length.

\begin{table}[t]
\centering
\caption{Input and output length statistics for evaluation datasets (ShareGPT and arXiv Summarization).}
\label{tab:dataset_stats}
\vskip 0.1in
\small
\resizebox{\columnwidth}{!}{%
\begin{tabular}{lccc|ccc}
\toprule
\textbf{Dataset} & \multicolumn{3}{c|}{\textbf{Input Length}} & \multicolumn{3}{c}{\textbf{Output Length}} \\
                 & \textbf{Mean} & \textbf{p90} & \textbf{Std}  & \textbf{Mean} & \textbf{p90} & \textbf{Std} \\
\midrule
ShareGPT             & 2340 & 5696  & 2088 & 438 & 834 & 265 \\
arXiv                & 9194 & 17152 & 5754 & 231 & 386 & 104 \\
\bottomrule
\end{tabular}}
\vspace{-0.12in}
\end{table}

\noindent\textbf{Traffic model:}
We define the request rate as the exogenous arrivals per second (req/s).
In our experiments, we generate arrivals with a Poisson process.

\noindent\textbf{SLOs:}
Following the convention of prior work~\cite{osdi-2024-sarathi, asplos-2025-podattention, isca-2025-windserve}, the TBT SLO is set to ${\sim}5\times$ the time to process 32 decode batches at 4096 tokens. The TTFT SLO is chosen separately for each model-dataset pair to reflect an appropriate operating point. All SLO settings are summarized in Table~\ref{tab:slo}. SLO attainment is evaluated per request: a request attains the SLO if its TTFT meets the TTFT SLO and, thereafter, the TBT of all generated tokens meets the TBT SLO.

\noindent\textbf{Energy:}
We measured the total energy consumed by the GPU devices during LLM serving using NVML’s API~\cite{nvidia-nvml, nsdi-2023-zeus}. Because NVML's API returns the cumulative energy used from the time the driver was loaded to the time of the call, we obtain a task’s energy by invoking it at the start and end of the task and taking the difference. Energy per token was measured as the total energy divided by the total number of prompt and generated tokens.

\begin{table}[t]
\centering
\caption{SLO constraints for the ShareGPT and arXiv datasets.}
\label{tab:slo}
\vskip 0.15in
\small
\begin{tabular}{lccc}
\toprule
\textbf{Model} & \textbf{Dataset} & \textbf{TTFT (s)} & \textbf{TBT (ms)} \\
\midrule
\texttt{Qwen}          & ShareGPT & 5  & 125 \\
\texttt{Qwen}          & arXiv    & 10 & 125 \\
\texttt{GPT}            & ShareGPT & 5  & 100 \\
\texttt{GPT}            & arXiv    & 10 & 100 \\
\bottomrule
\end{tabular}
\end{table}

\begin{figure}
    \centering
    \includegraphics[width=0.99\columnwidth]{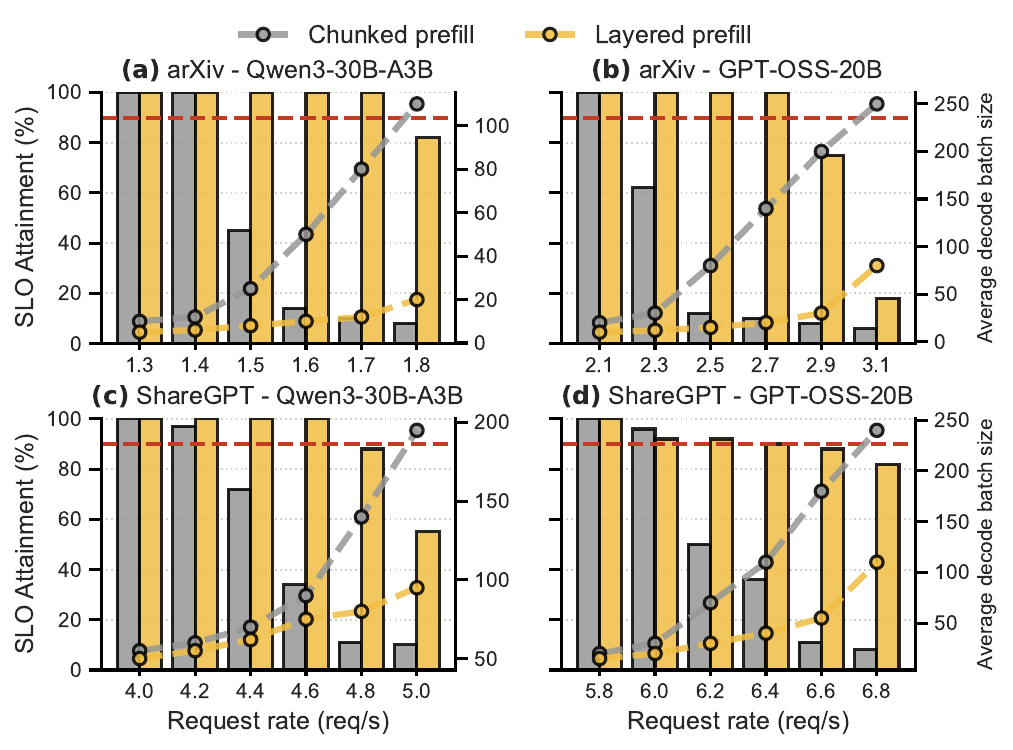}
    \vspace{-0.05in}
    \caption{SLO attainment under different request rates.
    The red horizontal line marks the effective SLO attainment threshold (90\%).}
    \label{fig:slo_distribution}
\end{figure}

\subsection{End-to-end SLO attainment}
\label{sec:e2e_slo}

Layered prefill delivers higher SLO attainment across models and workloads. Figure~\ref{fig:slo_distribution} compares chunked prefill and layered prefill across request rates for two models (\texttt{Qwen} and \texttt{GPT}) and two workloads (arXiv and ShareGPT).

\noindent\textbf{\texttt{Qwen}:}
(a) On arXiv, layered prefill sustains $\approx\!100\%$ SLO attainment through $1.7$\,req/s, while chunked prefill collapses at $1.5$\,req/s.
(c) On ShareGPT, layered prefill retains a clear advantage at $4.8$\,req/s whereas chunked prefill drops at $4.4$\,req/s.

\noindent\textbf{\texttt{GPT}:}
(b) On arXiv, layered prefill maintains $100\%$ SLO attainment across $2.1$--$2.7$\,req/s, whereas chunked prefill degrades quickly beyond $2.3$\,req/s.
(d) On ShareGPT, layered prefill also remains above $90\%$ across $5.8$--$6.4$\,req/s, while chunked prefill declines steeply at $6.2$\,req/s.

The dotted lines denote the average decode batch size. As shown, while maintaining SLO attainment above 90\%, the average decode batch size remains at most 32 for arXiv and 128 for ShareGPT.
According to Table~\ref{tab:moe_weight_ratio}, a decode batch size of 32 activates only about 55\% of experts, whereas a batch size of 128 activates roughly 86\%. Taken together, these observations indicate that layered prefill improves performance over chunked prefill by approximately 14--45\%, depending on the workload.
Moreover, comparing within the same model, the performance gap is larger on arXiv---where the prefill-to-decode length ratio is higher---than on ShareGPT. This suggests that as the number of chunks grows relative to the prompt length, chunking-induced degradation becomes more severe, and layered prefill mitigates this effect.
These trends are consistent with the detailed TTFT/TBT statistics in Table~\ref{tab:qwen30b_sharegpt_req1p3}, which show that layered prefill achieves both lower mean latency and tighter tail behavior compared to chunked prefill under the same workload conditions.

Finally, in chunked prefill, increasing the chunk size from $512$ to $1024$ reduced the number of chunks and lowered TTFT; however, the longer prefill time increased TBT and led to more frequent TBT violations, resulting in lower overall SLO attainment.

\begin{table}[t]
  \centering
  \small
  \caption{\texttt{Qwen} on arXiv (at 1.3\,req/s)}
  \label{tab:qwen30b_sharegpt_req1p3}
  \vskip 0.15in
  \small
  \begin{tabular}{lcccc}
    \toprule
    \textbf{Schedule} & \multicolumn{2}{c}{\textbf{TTFT (s)}} & \multicolumn{2}{c}{\textbf{TBT (ms)}} \\
                 & \textbf{Mean} & \textbf{p99} & \textbf{Mean}  & \textbf{p99} \\
    \midrule
    Chunked & 2.80 & 8.65 & 32.9 & 51.1 \\
    Layered & 1.24 & 4.10 & 21.5 & 37.1 \\
    \bottomrule
  \end{tabular}
\end{table}

\subsection{Breakdown of SLO Attainment}

\begin{figure}
    \centering
    \includegraphics[width=0.99\columnwidth]{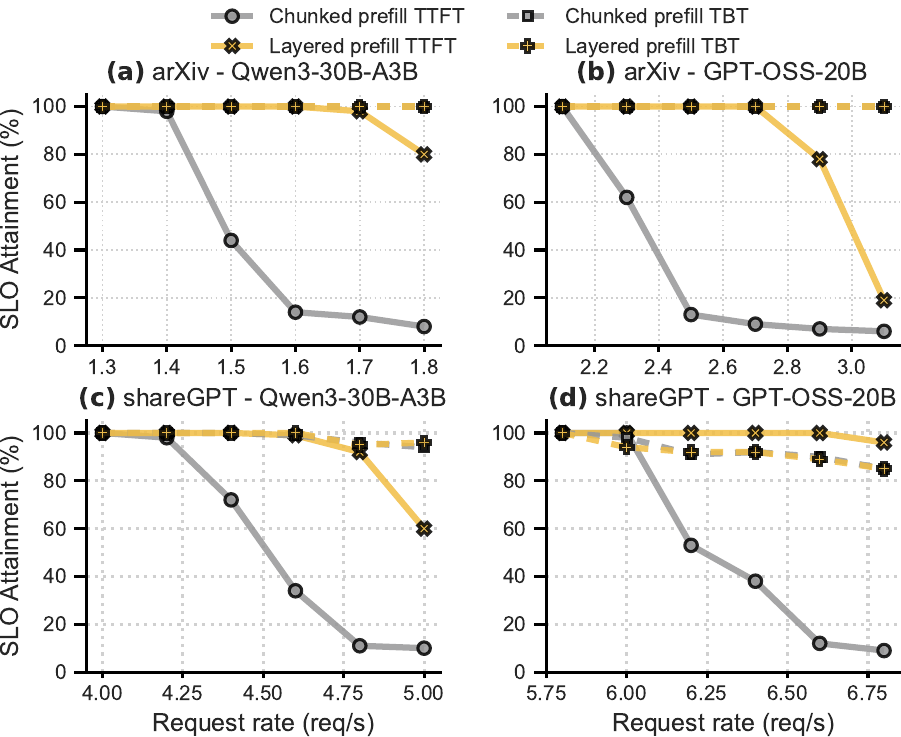}
    \vspace{-0.05in}
    \caption{Breakdown of SLO attainment by component across request rates.}
    \label{fig:slo_attainment}
\end{figure}

Layered prefill maintains higher overall SLO attainment mainly by preserving TTFT under load, while both schedulers already keep TBT near perfect.
Figure~\ref{fig:slo_attainment} decomposes end-to-end SLO attainment into its two components---TTFT and TBT---over increasing request rates, comparing chunked prefill to layered prefill. Across models and datasets, layered prefill sustains near-perfect TTFT attainment over a wider operating region than chunked prefill. As the system approaches saturation, layered prefill exhibits a more gradual decline in TTFT attainment, implying lower queuing delay and shorter prefill runtime relative to chunked prefill.

The TBT curves for both chunked prefill and layered prefill remain near 100\% across request rates, indicating that both schedulers effectively provide stall-free scheduling. Even as the request rate increases toward saturation, the TBT attainment exhibits little to no degradation, suggesting that both approaches can maintain stable TBT under severe load conditions. Exceptionally, when evaluating \texttt{GPT} on ShareGPT, high request rates inflate decode-batch size and trigger TBT violations in both chunked prefill and layered prefill. However, TTFT violations begin for chunked prefill at 6.2\,req/s, whereas layered prefill remains robust with no TTFT violation up to 6.6\,req/s.

\subsection{MoE Expert Load Traffic}

Layered prefill lowers expert-weight load traffic compared to chunked prefill, and the reduction is most significant on long-prompt workloads.
We quantified the memory pressure induced by expert activation using a simple counter:
the total number of \emph{expert weight load bytes} observed while processing a fixed trace of 100 requests.
A load byte is accumulated whenever an MoE expert’s parameters are brought into device memory for execution (either during prefill or decode).
Table~\ref{tab:expert_loads} reports this metric for the \texttt{Qwen} model on ShareGPT and arXiv under two schedulers:
chunked prefill with 512-token chunks and layered prefill.

\begin{table}[tb!]
\centering
\caption{Total expert weight loads for 100 requests on \texttt{Qwen}.}
\label{tab:expert_loads}
\vskip 0.1in
\small
\begin{tabular}{lccc}
\toprule
\textbf{Dataset} & \textbf{Scheduler} & \textbf{Total Loads} & \textbf{Reduction} \\
\midrule
\multirow{2}{*}{ShareGPT} & Chunked & 28.5 TB & - \\
                          & Layered & 25.1 TB & \textbf{-12.0\%} \\
\midrule
\multirow{2}{*}{arXiv}    & Chunked & 35.6 TB & - \\
                          & Layered & 21.7 TB & \textbf{-39.0\%} \\
\bottomrule
\end{tabular}
\end{table}

There are two key observations.
First, for ShareGPT, layered prefill reduces expert loads by \textbf{12\%}, consistent with fewer redundant reloads when prefill is distributed across layer groups and chunk-induced repetition is avoided.
Second, the effect is much larger on arXiv (\textbf{39\%} reduction), where long prompts would otherwise force many chunks, whereas layered prefill with large chunks preserves the sparsity benefit of MoE.
These reductions align with the SLO results in Figures~\ref{fig:slo_distribution} and \ref{fig:slo_attainment}: lower expert-load traffic correlates with higher SLO attainment at elevated request rates, particularly on workloads with high prefill-to-decode ratios.

\subsection{Token Generation Over Time}

\begin{figure}[t]
    \centering
    \includegraphics[width=0.82\columnwidth]{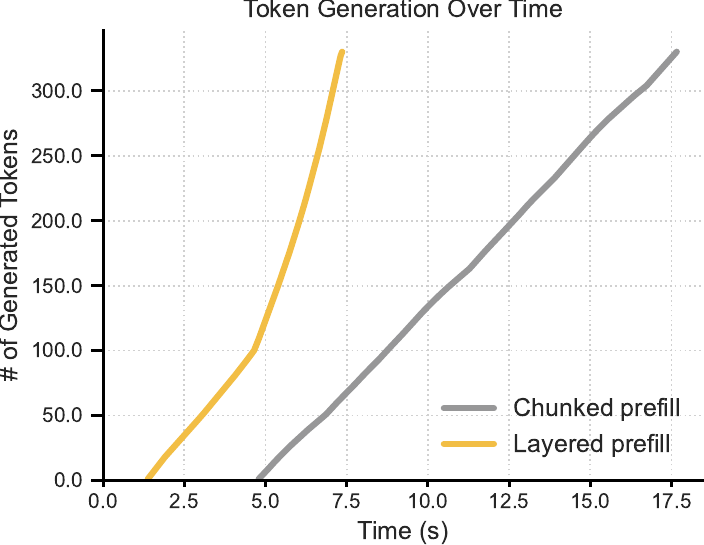}
    \vspace{-0.09in}
    \caption{Token generation over time on arXiv with \texttt{Qwen}.}
    \label{fig:token_generation}
\end{figure}

Layered prefill emits tokens earlier and maintains a higher generation rate than chunked prefill, yielding more tokens within the same wall-clock window.
In Figure~\ref{fig:token_generation}, we compared cumulative token output for a single request on \texttt{Qwen} using the arXiv workload at a request rate of 1.3\,req/s under chunked prefill and layered prefill.
The steeply rising middle interval reflects the period when layered prefill has quickly finished other requests’ prefills and runs in decode-only mode, so token generation accelerates.
These factors reduced the average end-to-end latency from \textbf{9.4} to \textbf{5.5}\,s (\textbf{-41\%}).
Overall, layered prefill reduces effective TTFT and increases throughput, which lowers end-to-end latency while preserving stall-free decoding.

\begin{table*}[t!]
\centering
\caption{Normalized energy on arXiv. We report energy per output token (mJ/tok; lower is better) together with mean/p99 TTFT and TBT at steady-state, SLO-compliant operating points. For each scheduler, the listed request rate is the highest that satisfies the latency SLO, so req/s reflects usable capacity.}
\label{tab:moe_energy}
\vskip 0.1in
\begin{tabular}{lccccccl}
\toprule
\multirow{2}{*}{\textbf{Model}} & \multirow{2}{*}{\textbf{Method}} & \multirow{2}{*}{\shortstack{Req. Rate\\(req/s)}} & \multicolumn{2}{c}{TTFT (s)} & \multicolumn{2}{c}{TBT (ms)} & \multirow{2}{*}{mJ / tok} \\
\cmidrule(lr){4-5} \cmidrule(lr){6-7}
 & & & Mean & p99 & Mean & p99 & \\
\midrule
\multirow{3}{*}{\texttt{Qwen}} & Chunked    & 1.3 & 2.80 & 8.65 & 32.2 & 51.3 & 56.6 \\
                      & Layered    & 1.3 & 1.24 & 4.10 & 21.4 & 37.5 & 51.7 (-9\%) \\
                      & Layered    & 1.6 & 2.46 & 8.01 & 28.1 & 41.2 & 44.2 (-22\%) \\
\cmidrule(lr){1-8}
\multirow{3}{*}{\texttt{GPT}} & Chunked    & 2.1 & 3.00 & 8.57 & 25.1 & 30.6 & 37.4 \\
                    & Layered    & 2.1 & 0.87 & 2.82 & 18.2 & 32.3 & 34.3 (-8\%) \\
                    & Layered    & 2.7 & 2.18 & 6.86 & 25.7 & 33.2 & 29.8 (-20\%) \\
\bottomrule
\end{tabular}
\end{table*}

\begin{table*}[ht]
\centering
\caption{Broader validation across GPUs, models, and workloads.}
\label{tab:broader_validation}
\vskip 0.1in
\small
\begin{tabular}{lllcccc}
\toprule
\textbf{GPU} & \textbf{Model} & \textbf{Method} & \multicolumn{2}{c}{\textbf{TTFT (s)}} & \multicolumn{2}{c}{\textbf{TBT (ms)}} \\
& & & \textbf{Mean} & \textbf{p99} & \textbf{Mean} & \textbf{p99} \\
\midrule
A100x2 & Qwen3-30B-A3B & Chunked (512) & 1.32\textcolor{white}{0} & \textcolor{white}{0}5.40\textcolor{white}{0} & 22.4 & 70.4 \\
& (arXiv, 0.5 req/s) & Layered ($N_{lg}=16$) & 0.962 & \textcolor{white}{0}3.81\textcolor{white}{0} & 19.3 & 53.6 \\
\midrule
H100x2 & gpt-oss-120b (fp4) & Chunked (512) & 3.11\textcolor{white}{0} & \textcolor{white}{0}8.82\textcolor{white}{0} & 34.9 & 47.2 \\
& (arXiv, 1.2 req/s) & Layered ($N_{lg}=12$) & 1.09\textcolor{white}{0} & \textcolor{white}{0}3.89\textcolor{white}{0} & 20.5 & 44.8 \\
\midrule
H100x8 & Qwen3-235B-A22B & Chunked (512) & 4.25\textcolor{white}{0} & 11.6\textcolor{white}{00} & 41.7 & 50.5 \\
& (arXiv, 1.2 req/s) & Layered ($N_{lg}=16$) & 2.26\textcolor{white}{0} & \textcolor{white}{0}7.42\textcolor{white}{0} & 33.1 & 45.0 \\
\midrule
H100x8 & Qwen3-235B-A23B & Chunked (512) & 6.65\textcolor{white}{0} & 16.4\textcolor{white}{00} & 50.2 & 70.2 \\
& (ShareGPT, 3.5 req/s) & Layered ($N_{lg}=16$) & 3.35\textcolor{white}{0} & \textcolor{white}{0}8.48\textcolor{white}{0} & 50.8 & 88.5 \\
\midrule
H100x8 & gpt-oss-120b (fp4) & Chunked (512) & 1.05\textcolor{white}{0} & \textcolor{white}{0}3.45\textcolor{white}{0} & 14.4 & 18.6 \\
& (arXiv, 3.0 req/s) & Layered ($N_{lg}=12$) & 0.530 & \textcolor{white}{0}1.75\textcolor{white}{0} & 11.8 & 19.3 \\
\midrule
H100x8 & gpt-oss-120b (fp4) & Chunked (512) & 0.212 & \textcolor{white}{0}0.791 & 12.2 & 19.6 \\
& (ShareGPT, 6.5 req/s) & Layered ($N_{lg}=12$) & 0.178 & \textcolor{white}{0}0.572 & 11.3 & 21.0 \\
\bottomrule
\end{tabular}
\end{table*}

\subsection{Energy per Output Token}
\label{sec:energy_arxiv_norm}

Table~\ref{tab:moe_energy} reports \emph{energy per output token} (mJ/tok) together with latency metrics (mean/p99 TTFT and TBT) at steady-state operating points on arXiv.
We selected for each scheduler the highest request rate that still satisfies our latency SLO targets (TTFT and TBT), so differences in req/s reflect usable capacity under the same QoS envelope.

\noindent\textbf{\texttt{Qwen}:}
Relative to chunked prefill at 1.3~req/s, layered prefill sustains 1.6~req/s ($+23\%$) while reducing energy per token from \textbf{56.6} to \textbf{44.2}\,mJ/tok ($\mathbf{-22\%}$).
Latency QoS is maintained or improved (TTFT decreasing; TBT comparable within the SLO).
Even at the same request rate, we observed a 9\% reduction in energy per token. In this setting, mean TTFT drops by more than \textbf{50\%} compared to chunked prefill at the same rate.

\noindent\textbf{\texttt{GPT}:}
Relative to chunked prefill at 2.1\,req/s, layered prefill sustains 2.7\,req/s ($+29\%$) while reducing energy per token from \textbf{37.4} to \textbf{29.8}\,mJ/tok ($\mathbf{-20\%}$).
Latency remains within the SLO envelope, with improved TTFT and comparable TBT.
When the request rate is the same, energy per token decreases by 8\% as well.
Under the same request rate, mean TTFT decreases by more than \textbf{70\%} over chunked prefill.

\noindent\textbf{Takeaways:}
Across both models, layered prefill delivers \textbf{20--22\%} lower energy per token at higher sustainable request rates, with equal-or-better TTFT and comparable TBT.
These normalized gains are consistent with our microbenchmarks: avoiding repeated expert-weight reloads reduces HBM/NVLink traffic in MoE layers, translating into lower per-token energy without sacrificing latency.

\subsection{Broader Validation Across GPUs and Models}
To demonstrate the robustness of layered prefill, we evaluated it across different GPUs (A100 and H100), model scales (30B to 235B), and precision regimes (including 4-bit fp4 weights). Table~\ref{tab:broader_validation} summarizes the results. Across all settings, layered prefill 
reduces TTFT while maintaining or improving TBT compared to chunked prefill. This confirms that the key benefit of layered prefill is robust across hardware generations, model scales, and precision regimes.

\subsection{Comparison to Prefill/Decode Disaggregation}
We compared layered prefill against prefill/decode disaggregation on \texttt{Qwen} using an H100x2 server. For disaggregation, one GPU was assigned to prefill and the other to decode, whereas chunked and layered prefill used both GPUs via tensor parallelism. As shown in Table~\ref{tab:disaggregation}, layered prefill achieves significantly better TTFT than disaggregation, while disaggregation achieves lower TBT due to dedicated decode resources. Layered prefill improves both TTFT and TBT over chunked prefill, providing a more balanced trade-off without requiring explicit system-level disaggregation and dynamic capacity balancing.

\begin{table}[t!]
\centering
\caption{Comparison with disaggregated serving.}
\label{tab:disaggregation}
\vskip 0.1in
\small
\begin{tabular}{lcccc}
\toprule
\textbf{Method} & \multicolumn{2}{c}{\textbf{TTFT (s)}} & \multicolumn{2}{c}{\textbf{TBT (ms)}} \\
& \textbf{Mean} & \textbf{p99} & \textbf{Mean} & \textbf{p99} \\
\midrule
Chunked prefill & 3.00 & \textcolor{white}{0}9.15 & 32.1 & 49.3 \\
Disaggregated & 3.94 & 11.2\textcolor{white}{0} & 16.4 & 25.8 \\
Layered prefill & 1.24 & \textcolor{white}{0}4.59 & 19.8 & 35.9 \\
\bottomrule
\end{tabular}
\end{table}

\subsection{Relaxed TBT SLO and Throughput Tradeoff}
Under a relaxed TBT SLO, chunked prefill can use larger chunks and layered prefill can use fewer layer groups. As the SLO becomes looser, both saturate and converge, since layered prefill mainly reduces the per-chunk overhead of chunked prefill, which becomes negligible at large chunk sizes. On \texttt{Qwen} (H100x2, arXiv, 2.5 req/s), chunked prefill with chunk size 2048 and layered prefill with $N_{lg}=4$ achieve nearly identical TBT and TTFT (Mean TTFT $\approx 2.47$s, Mean TBT $\approx 78.6$ms).

We also compared peak offline throughput on H100x2 (\texttt{Qwen}, arXiv). With chunk size 512 for chunked prefill and $N_{lg}=16$ for layered prefill, layered prefill achieves higher throughput (391.59 vs. 315.36 tokens/sec). This is because chunked prefill can still under-utilize MoE experts despite executing all layers each step, whereas layered prefill improves utilization for the active layers and reduces unnecessary expert weight loading. The throughput gap narrows with larger chunk sizes (and smaller $N_{lg}$), since redundant MoE loading becomes less significant.

\subsection{Experiments on Group Size $N_{lg}$}
Layered prefill introduces the layer group size $N_{lg}$ as a tuning knob. Increasing $N_{lg}$ increases TTFT but improves TBT, while decreasing $N_{lg}$ reduces TTFT at the cost of higher TBT. Table~\ref{tab:group_size} shows this trade-off on H100x2 with \texttt{Qwen} (arXiv, 1.4 req/s), demonstrating that $N_{lg}$ provides a practical control knob to match different latency SLOs.

\begin{table}[tb!]
\centering
\caption{Ablation on group size $N_{lg}$.}
\label{tab:group_size}
\vskip 0.1in
\small
\begin{tabular}{ccccc}
\toprule
\textbf{Group size} & \multicolumn{2}{c}{\textbf{TTFT (s)}} & \multicolumn{2}{c}{\textbf{TBT (ms)}} \\
($N_{lg}$) & \textbf{Mean} & \textbf{p95} & \textbf{Mean} & \textbf{p99} \\
\midrule
2 & 0.480 & 1.62 & 20.9 & 138 \\
4 & 0.566 & 2.01 & 20.6 & 84.5 \\
8 & 0.768 & 2.81 & 20.8 & 54.6 \\
16 & 1.27\textcolor{white}{0} & 3.47 & 19.7 & 35.7 \\
\bottomrule
\end{tabular}
\end{table}

\subsection{Long-Context Discussion}
As context length grows, attention becomes more expensive, which can reduce the relative impact of MoE weight traffic. However, our profiling shows that MoE remains a substantial fraction of runtime even at long context lengths (e.g., 32K). At chunk size 512, MoE accounts for 57.7\% of runtime at 8K input and 44.9\% at 32K input. We validated this with $\sim$30K-context ShareGPT requests on \texttt{Qwen} (H100x2, 0.1 req/s). Layered prefill ($N_{lg}=16$) continues to reduce TTFT substantially (1.74s vs. 2.91s) while slightly improving TBT (14.9ms vs. 15.6ms) compared to chunked prefill (chunk=512).

\subsection{Responsiveness, Smoothness, and Fairness}
Perceived responsiveness is not fully captured by average TTFT/TBT alone. We evaluated token smoothness (TBT std) and a fairness metric based on Jain's index over per-request decode throughput. Table~\ref{tab:smoothness} summarizes results under an online workload (H100x2, arXiv, 1.4 req/s). Layered prefill substantially improves TTFT and reduces both mean and tail TBT, while also improving smoothness (lower TBT std). Importantly, Jain's fairness index also increases, suggesting that layered prefill does not introduce additional per-request unfairness in streaming throughput. We also tested bursty arrivals (Coefficient of Variation = 1.83) and observed the same trend, with layered prefill mitigating tail amplification.

\begin{table}[tb!]
\centering
\caption{Smoothness and fairness metrics.}
\label{tab:smoothness}
\vskip 0.1in
\small
\setlength{\tabcolsep}{5pt}
\begin{tabular}{lcccccc}
\toprule
\textbf{Method} & \multicolumn{2}{c}{\textbf{TTFT (s)}} & \multicolumn{2}{c}{\textbf{TBT (ms)}} & \textbf{TBT std} & \textbf{Jain} \\
& \textbf{Mean} & \textbf{p99} & \textbf{Mean} & \textbf{p99} & \textbf{(ms)} & \textbf{Index} \\
\midrule
\multicolumn{7}{l}{\textit{Poisson Arrivals}} \\
Chunked & 3.00 & \textcolor{white}{0}9.15 & 32.1 & 49.3 & 10.7 & 0.835 \\
Layered & 1.24 & \textcolor{white}{0}4.59 & 19.8 & 35.9 & 8.51 & 0.882 \\
\midrule
\multicolumn{7}{l}{\textit{Bursty Arrivals (CV=1.83)}} \\
Chunked & 5.63 & 16.7\textcolor{white}{0} & 31.2 & 48.6 & 11.2 & 0.809 \\
Layered & 2.55 & \textcolor{white}{0}8.07 & 22.3 & 36.6 & 8.89 & 0.850 \\
\bottomrule
\end{tabular}
\end{table}

\subsection{Ablation on a Non-MoE Model}
We evaluated a dense (non-MoE) model to test whether layered prefill helps beyond MoE. On Qwen3-8B with a single H100 (arXiv, 1.5 req/s), layered prefill is slower than chunked prefill (Mean TTFT 1.45s vs. 0.955s, Mean TBT 20.3ms vs. 16.2ms). This is expected because chunked prefill already sustains high GPU utilization in dense models, while layered prefill can reduce utilization without MoE-specific reuse benefits. Thus, layered prefill is most effective for MoE models where expert activation is sparse and strict TBT SLOs require small chunk sizes.

\section{Related Work}

\subsection{LLM Serving Systems}
The rapid proliferation of large language models has spurred extensive work on efficient serving infrastructures.
Early systems such as FasterTransformer~\cite{github-2019-fasttransformer} and DeepSpeed-Inference~\cite{icml-2022-deepspeed_moe} introduced kernel-level optimizations (e.g., fused attention and tensor parallelism) to accelerate single-model inference.
Recent work has focused on improving multi-user serving efficiency under resource contention.
For example, vLLM~\cite{sosp-2023-vllm} proposes PagedAttention to manage the KV cache with fine-grained memory virtualization, enabling high-throughput decoding with large batches.
Sarathi-Serve~\cite{osdi-2024-sarathi} introduces adaptive batching and request scheduling to balance latency and throughput.
Our work builds on this line by revisiting the scheduling of prefill workloads, specifically addressing inefficiencies that arise in MoE models under chunked prefill.

\subsection{Energy Efficiency of LLMs}
Beyond raw performance, energy efficiency has emerged as a critical concern in large-scale deployment.
Prior studies~\cite{arxiv-2024-memory, frontier-2025-flexnpu} report that memory traffic dominates the energy footprint of modern accelerators, motivating techniques that reduce parameter movement or KV cache access.
Architectural proposals such as MLA~\cite{arxiv-2024-deepseek_v2} compress KV caches to lower both memory capacity and power demand.
Serving-level approaches, including early-exit strategies and speculative decoding, have been analyzed for their potential to cut energy per token.
However, most prior work evaluates dense transformer models; relatively little attention has been given to the unique energy behavior of MoE models, where expert weight loading can significantly amplify off-chip memory energy consumption.
Our work complements this line by quantifying the energy overhead of MoE under different prefill scheduling policies and by demonstrating that layered prefill can reduce both latency and energy costs in long-context workloads.

\subsection{Scheduling and Disaggregation}
FlexGen~\cite{icml-2023-flexgen} targets throughput-oriented scenarios with very loose SLOs, focusing on layer-wise offloading and batching strategies (e.g., zig-zag schedule) to maximize efficiency over large batches under memory pressure. In contrast, layered prefill is designed for online serving under strict streaming latency requirements, specifically targeting the MoE expert reuse failure mode and per-chunk overhead induced by chunked prefill under tight TBT constraints. While both approaches share the high-level idea of changing scheduling granularity, they operate under fundamentally different constraints and bottlenecks.

Disaggregated serving systems, such as Dynamo~\cite{nvidia-dynamo} and DistServe~\cite{osdi-2024-distserve}, change the placement and pipeline structure across engines or nodes by separating prefill and decode. Layered prefill is orthogonal to these systems because it changes the scheduling unit inside a single engine. These techniques can be combined, and layered prefill can improve expert reuse even within a colocated or partially disaggregated setup.

Layered prefill is compatible with speculative decoding~\cite{icml-2023-speculative}, which restructures the decode process into draft/verify cycles. By treating verification as a regular decode batch, layered prefill can be used together with speculative decoding.

\section{Conclusion}

Conventional chunked prefill scheduling is inefficient for MoE models, as it amplifies redundant expert weight loads, increasing both latency and energy consumption.
This paper introduces \emph{layered prefill}, a new scheduling strategy that partitions a model along the layer dimension rather than the token dimension.
This approach fundamentally eliminates the redundant memory traffic caused by chunking.
Our evaluation on representative MoE models shows that layered prefill consistently improves SLO attainment, reduces expert-load traffic by up to \textbf{39\%}, and lowers TTFT by up to \textbf{70\%}, end-to-end latency by \textbf{41\%}, and per-token energy by up to \textbf{22\%} on long-context workloads.
These results underscore the importance of co-designing scheduling policies with emerging model architectures.

Future work can extend this approach to complex multi-GPU environments and explore adaptive layer grouping strategies, particularly for models where the layer count is not an even multiple of the group count.
Further research could also jointly optimize for throughput, latency, and energy efficiency at a data-center scale.

\section*{Acknowledgements}
This work was supported by Mobile eXperience (MX) Business, Samsung Electronics Co., Ltd.
It was also supported by the Institute of Information \& Communications Technology Planning \& Evaluation (IITP) grant funded by MSIT (RS-2021-II211343, IITP-2026-RS-2023-00256081).
Our source code is available at \url{https://github.com/scale-snu/layered-prefill}


\bibliography{refs}
\bibliographystyle{mlsys2026}

\appendix



\end{document}